\documentclass[runningheads]{llncs}
\usepackage{listings}
\usepackage{booktabs}
\usepackage{tabularx}
\usepackage{array}
\usepackage[T1]{fontenc}
\usepackage{seqsplit}
\usepackage{xcolor}
\newcommand{\rev}[1]{#1}
\usepackage{xurl}
\usepackage{hyperref}
\usepackage{graphicx}
\begin{document}
\title{Authoring and Management of Transparent Research Integrity Assessments of Randomised Clinical Trial Publications Using LLM-assisted Tools and Provenance Knowledge Graphs }
%
%

\author{
Milan Markovic\inst{1,2}\orcidID{0000-0002-5477-287X} \and
Goutham Indukuri\inst{2}\orcidID{0009-0005-2520-5007} \and
Somayajulu Sripada\inst{2}\orcidID{0000-0002-5428-8383} \and
Colby J. Vorland\inst{3}\orcidID{0000-0003-4225-372X} \and
Jack Wilkinson\inst{4}\orcidID{0000-0003-3513-4677} \and
Clare Robertson\inst{5}\orcidID{0000-0001-6019-6795} \and
Mark Bolland\inst{6}\orcidID{0000-0003-0465-2674} \and
Andrew Grey\inst{6}\orcidID{0000-0002-7803-0096} \and
Miriam Brazzelli\inst{5}\orcidID{0000-0002-7576-6751} \and
Alison Avenell\inst{5}\orcidID{0000-0003-4813-5628}
}

\authorrunning{Markovic et al.}
%
\institute{
Interdisciplinary Institute, University of Aberdeen, UK\\
\email{milan.markovic@abdn.ac.uk}
\and
Department of Computing Science, University of Aberdeen, UK\\
\email{\{goutham.indukuri, yaji.sripada\}@abdn.ac.uk}
\and
Department of Epidemiology and Biostatistics, Indiana University, USA\\
\email{cvorland@iu.edu}
\and
Centre for Biostatistics, University of Manchester, UK\\
\email{jack.wilkinson@manchester.ac.uk}
\and
Aberdeen Centre for Evaluation, University of Aberdeen, UK \\
\email{\{c.robertson, m.brazzelli, a.avenell\}@abdn.ac.uk}
\and
Department of Medicine, University of Auckland, NZ \\
\email{\{a.grey, m.bolland\}@auckland.ac.nz}
}

\maketitle              
\begin{abstract}

Systematic reviews of Randomised Controlled Trials (RCTs) are routinely used as evidence for clinical care guidelines. Such evidence has to meet high research integrity standards to prevent low quality or false research outputs influencing the clinical care. However, assessing research integrity of published RCTs is a complex process requiring manual effort, and potentially resulting in diverse opinions of the human assessors.  This paper describes INSPECT-AI, an LLM-based interactive tool that assists human reviewers with research integrity assessments of published RCTs based on the community approved INSPECT-SR framework, and the Research Integrity Provenance and Evidence ontology (RIPE-O) for documenting the provenance of the assessment process. In addition, we present the Research Integrity Provenance and Evidence knowledge graph (RIPE-KG), an initial set of 140 expert research integrity assessments of 95 RCT publications  generated by INSPECT-AI and described using RIPE-O.   

 \textbf{Permanent Resource Identifier:} \url{https://w3id.org/ripe}

\end{abstract}
\section{Introduction}
The field of metascience, which involves the study of scientific practice has experienced growing interest in recent years, particularly due to challenges related to reproducibility and the academic integrity of scientific outputs \cite{peterson2023metascience}. 
The UK Research Integrity Office sets out five principles of research integrity including honesty, rigor, transparency and communication, care and respect, and accountability \cite{UKRIO2025}.  Problematic research publications (e.g., those containing falsified or misinterpreted data, incorrect statistical analysis) have a significant impact on the health care domain. For example, systematic reviews of Randomised Controlled trials (RCTs) are used by the National Institute for Health and Care Excellence (NICE),\footnote{https://www.nice.org.uk/} National Institute for Health and Care Research (NIHR) Evidence Synthesis Programme,\footnote{https://www.nihr.ac.uk/funding-programmes/evidence-synthesis} the Scottish Intercollegiate Guidelines Network (SIGN)\footnote{https://www.sign.ac.uk/} and the Cochrane Library\footnote{https://www.cochranelibrary.com/} as the best evidence to inform clinical care. Prior work by members of our research team demonstrated how 27 RCTs with integrity concerns were incorporated into 88 systematic reviews or clinical guidelines, changing findings in over half - 87\% with a substantial change in the direction of effect \cite{avenell2024randomized}. Cochrane reviewers applying an integrity assessment checklist removed 25\% of trials, requiring one third of systematic reviews' findings to be updated \cite{weeks2023trustworthiness}.  However, manual completion of checklists and statistical tools is time-consuming, which prevents the scalability of their application. For example, reviewers have to manually search for retraction notices or any relevant comments, assess the co-authors profiles by inspecting their previous publications, and compare information from the publication with other external information.  For this reason, we have developed the INSPECT-AI tool which utilises Large Language Models (LLMs) to increase the efficiency of the human evaluator following the INSPECT-SR assessment guidelines \cite{Wilkinson2025.09.03.25334905} - manual research integrity guidance endorsed by the Cochrane Collaboration~\cite{cochrane_methods}. \rev{In line with INSPECT-SR, the INSPECT-AI tool focuses on the honesty principle in research publishing by assessing whether the data and findings reported in RCT publications are sufficiently trustworthy for use in evidence synthesis.}

However, even with such a tool, research integrity assessments are inherently susceptible to variability, as evaluators following the same guidelines may employ subjective judgements. Therefore, we argue that FAIR principles \cite{wilkinson2016fair} should apply to the publication of provenance traces of such assessments, providing the standardised infrastructure needed to make their results transparent, reproducible, and reusable across the scientific community. This motivated us to create the Research Integrity Provenance and Evidence (RIPE) Observatory,~\footnote{ \url{https://w3id.org/ripe}} a collection of semantic resources for capturing and publishing provenance traces of research integrity assessments in a knowledge graph. 

In this paper, we present the initial components of the RIPE Observatory including:

\begin{itemize}
    \item INSPECT-AI, an AI tool enabling researchers to perform research integrity assessments and to generate semantic provenance traces documenting the assessment processes and their outcomes.\newline
    \textbf{Permanent Identifier:} \url{https://w3id.org/ripe/inspect-ai}
    \item Research Integrity Provenance and Evidence Ontology (RIPE-O) for documenting the provenance of the integrity assessment process.\newline
    \textbf{Permanent Identifier:} \url{https://w3id.org/ripe/ripe-o}
    \item Research Integrity Provenance and Evidence Knowledge Graph (RIPE-KG) with the web-based GUI, containing at the time of writing provenance traces describing 140 integrity assessments of 95 publications. 
    \newline\textbf{Permanent Identifier:} \url{https://w3id.org/ripe/ripe-kg}
\end{itemize}

\rev{While the core semantic contributions of this paper are RIPE-O and RIPE-KG, the INSPECT-AI assessment tool is included because it informed the development of RIPE-O and completes the proposed pipeline. However, we also envision RIPE-O as an ontology for integrating outputs from other existing and future research integrity assessment tools into RIPE-KG or similar knowledge graphs. }

\section{Related Work}

Knowledge graph technologies have been successfully used to model, integrate, and manage large volumes of scholarly data from heterogeneous sources. For example, OpenAIRE Research Graph integrates information on various research outputs (e.g., datasets, publications, software) including metadata from publishers, repositories, funders, and research infrastructures \cite{manghi2019openaire}. Similarly, Scholia \cite{nielsen2017scholia} uses Wikidata to generate semantic scholarly profiles for researchers, organizations, journals, and research topics via the SPARQL Query Service. Semantic Scholar \cite{kinney2023semantic}, Open Research Knowledge Graph \cite{jaradeh2019open}, SemOpenAlex \cite{farber2023semopenalex} also integrate bibliometric data from a range of proprietary and public resources. In addition, such platforms enhance their graphs with results of various crowdsourced and AI-supported analysis of scholarly data such as topics and concepts extracted from publications to enable further analysis and filtering \cite{verma2023scholarly}. However, to the best of our knowledge, none of the existing scholarly knowledge graphs integrate data related to \rev{publication trustworthiness or other research integrity assessments.} 

To date, discovering problematic RCTs remains a highly manual effort. Research integrity ‘sleuths’ may alert journals and publishers to untrustworthy research publications \cite{retractionwatch2018sleuths}, however, publishers can be very slow to retract articles or post expressions of concern on their webpages, if they do at all. A high proportion of untrustworthy publications remain unflagged by journals and publishers years after notification \cite{grey2025ten,grey2025inconsistency}. \rev{Other tools supporting research integrity assessments of published research include the TRACT~\cite{mol2023tract}, RIA~\cite{weibel2023ria}, CPC-TST~\cite{weeks2023trustworthiness}, and REAPPRAISED~\cite{grey2020reappraised} checklists, alongside early experimentation with semi-automating the TRACT checklist using GPT-4o \cite{au2025tract}. However, these mostly manual tools are not endorsed by Cochrane and do not produce semantic data.} The INSPECT-SR checklist \footnote{\url{https://inspect.sr/}} was developed by international integrity researchers through a consensus and prototype testing process \cite{Wilkinson2025.09.03.25334905}. The 21 checks cover four domains: inspecting post-publication notices; inspecting reported conduct, governance and transparency; inspecting text and figures (for image manipulation and text duplication); and inspecting results, which mainly relate to numerical data and statistical results. This checklist is endorsed by the Cochrane Collaboration~\cite{cochrane_methods} which is an international organisation of researchers, patients and carers that produces systematic reviews of healthcare and public health interventions to inform international clinical guidelines. The work presented in this paper focuses on automating part of the INSPECT-SR checklist.

\section{Methodology}

INSPECT-AI, RIPE-O and RIPE-KG resulted from interdisciplinary efforts of systematic reviewers, clinicians,  statisticians, and computer scientists. \rev{INSPECT-AI was developed first to support INSPECT-SR assessments, and RIPE-O and RIPE-KG followed to document and publish the resulting assessment provenance.} Knowledge transfer and shared understanding were key requirements of success as we  observed that real, semi-functional prototypes resembling the final product were more effective than paper prototypes or detailed technical explanations. We have therefore adopted an agile software development methodology and produced a number of early iterative software prototypes and demos. This enabled the non-technical members of the team to better understand the challenges and limitations associated with AI technologies in this context. Such enhanced understanding influenced, for example, the decision to focus on a human-in-the-loop solution instead of a fully automated AI application. Although the requirements for integrity assessments were clearly defined in the INSPECT-SR guidelines, they were at times difficult to understand for computer scientists \rev{due to a lack of domain knowledge}. This was mitigated by providing a non-trivial set of 50 diverse examples of existing real-world problematic publications and detailed guidance on the assessment process by biomedical researchers. Early prototyping of end-to-end solutions rather than isolated components also enabled engagement with several evidence synthesis groups who were able to test the application in a real operational setting. Because the main initial objective of the project was to create an AI-powered app that would allow users to perform INSPECT-SR assessments, the semantic data model did not influence the design of the application. Instead, semantic mechanisms focused on capturing the provenance of the decision-making process enabled by the app (i.e., the research integrity assessment). Provenance graphs were generated by applying YARRRML mappings~\cite{heyvaert2018declarative} to INSPECT-AI process logs.

To develop RIPE-O, we followed the four-stage Linked Open Terms methodology (LOT)~\cite{POVEDAVILLALON2022104755}. As our intention with RIPE-O was to document the provenance of the integrity assessment process, we first formed the competency questions guided by the selected INSPECT-SR checks and framed by the 7W's of Provenance (Who, What, Where, Which, Why, When, How)~\cite{10.5555/2889875.2889882}. Building an early prototype of a web interface displaying the potential provenance data generated during the assessment process allowed us to engage non-technical team members to validate and refine the proposed ontology scope. RIPE-O extends the Threat Intelligence Decision Ontology (TIDO)~\cite{roothaert2025tido}, which adopts a forensic analysis approach to documenting provenance of decision-making processes and aligns with the PROV-O ontology for documenting provenance traces. SPAR ontologies including FABIO and CITO~\cite{peroni2018spar} were reused to model additional bibliographic information of research outputs considered in the research integrity assessments. We used Prot{\'e}g{\'e}~\cite{protege} to develop the ontology and WIDOCO~\cite{garijo2017widoco} for the documentation. Permanent URLs for the ontology were handled through the w3id platform%
,\footnote{\url{https://github.com/perma-id/w3id.org/\#permanent-identifiers-for-the-web}} %
 enabling content negotiation in multiple serialisations and the maintenance of the semantic artefacts is supported via GitHub.

\begin{figure}[h]
\centering
\includegraphics[width=\textwidth]{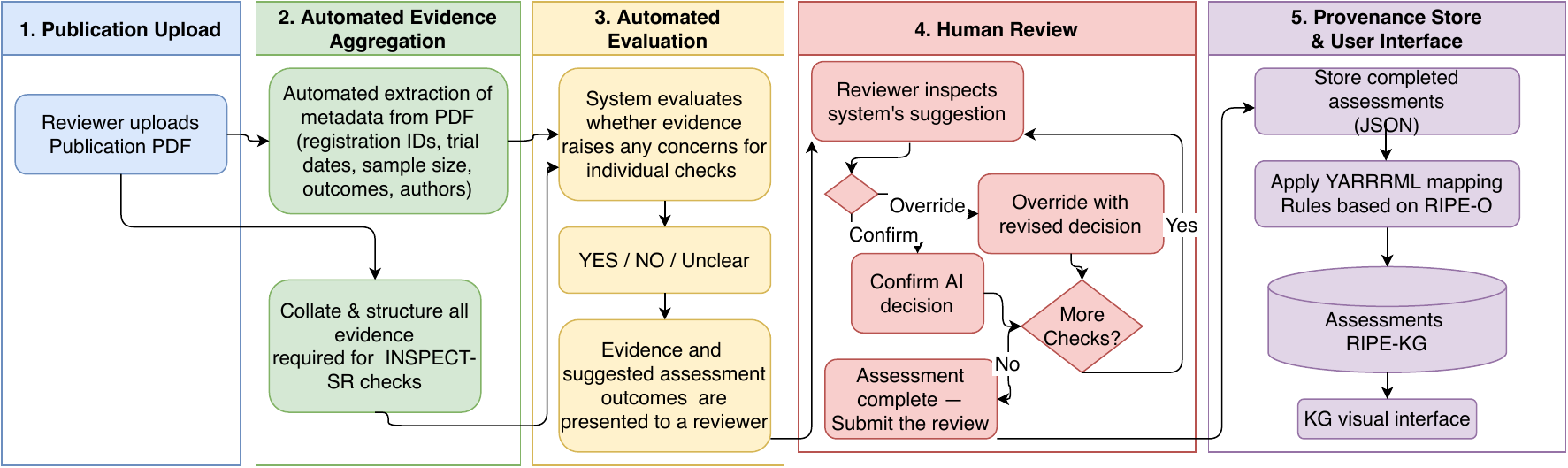}
\caption{The INSPECT-AI workflow detailing two automated steps for evidence aggregation and initial evaluation, human review step, and finally storage and publishing of provenance traces documenting the assessment process.}
\label{fig:workflow}
\end{figure}

\section{INSPECT-AI Tool}

To produce a systematic review, reviewers have to assess whether an RCT publication should be included. One of the criteria preventing inclusion of otherwise relevant publication is the  potential issue related to research integrity (e.g., concerns raised by the community, retraction notices). \rev{INSPECT-AI is designed to support research integrity assessment by semi-automating INSPECT-SR checks of RCT trustworthiness and documenting the provenance trace describing how the assessment was performed (Figure \ref{fig:workflow}).} After the reviewer uploads a PDF of the publication, INSPECT-AI  extracts structured metadata using GROBID~\cite{grobid}, including the publication DOI, title, authors, and reference list. In parallel, INSPECT-AI uses LLM (GEMINI 2.0 Flash\footnote{https://ai.google.dev/gemini-api/docs/models/gemini-2.0-flash})  to extract additional information such as registration identifiers and the study timeline (i.e., start and end of the recruitment, trial dates) from the publication's text. Each piece of LLM-generated evidence is accompanied by a short explanation of why particular values were extracted. This is important in a human-in-the-loop setting because reviewers can later  quickly identify and verify the extracted information within the publication. \rev{Prompts, response validation and testing are documented in the public LLM guide.}\footnote{\url{https://github.com/RIPE-Observatory/INSPECT-AI\_open\_source/blob/v1.0.0/LLM\_GUIDE.md}} \rev{The LLM outputs are treated as suggestions that can be corrected or overridden by reviewers before the final assessment is recorded.} Extracted metadata such as DOI and title, are  used to query external data sources, including the Retraction Watch Database~\cite{retractionWatchDatabase}, and PubPeer~\cite{pubpeer} for external evidence linked to the publication. When trial registration identifiers are extracted from the PDF, they are used to retrieve registration-related evidence from the ClinicalTrials.gov API~\cite{ClinicalTrialsgov} and the WHO ICTRP Search Portal~\cite{whoictrp}. 

\begin{figure}[htbp]
      \centering
      \includegraphics[width=\textwidth]{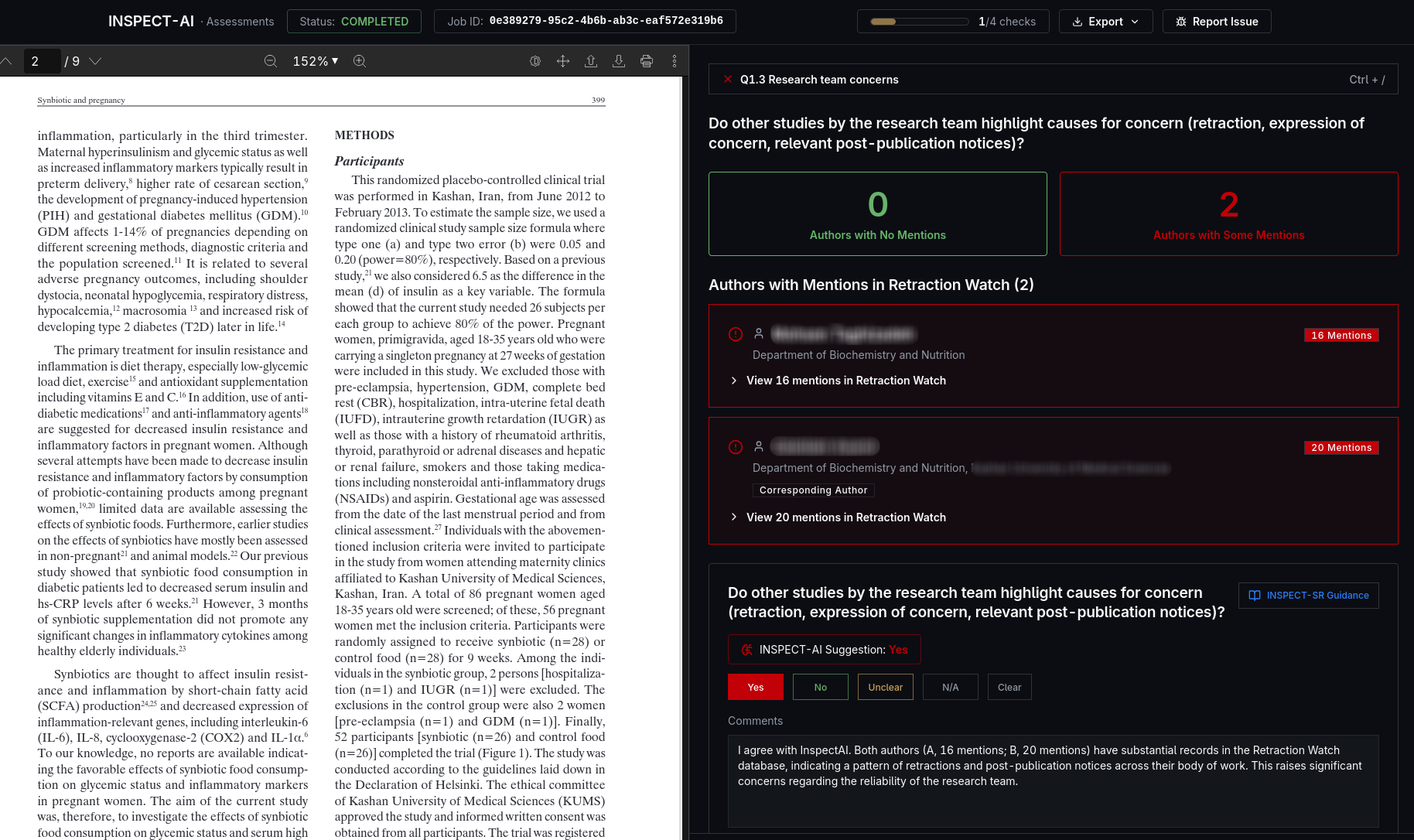}
      \caption{Web interface of INSPECT-AI application. The assessed publication is presented on the left. Evidence and the results corresponding to individual checks are displayed on the right side of the UI.}
      \label{fig:inspectAI}
\end{figure}

Following the evidence gathering step, each evidence piece is evaluated against the relevant INSPECT-SR check using conditional logic with pre-defined rules, and the system generates suggested outcomes stating whether concerns have been found, or whether the evidence is unclear. For example, in the prospective registration check, the recruitment start date extracted from the publication text is compared with the registration date retrieved from the trial registry. If the registration date is after the recruitment start date, the system suggests that concerns relating to retrospective trial registration have been found.
The INSPECT-AI web interface (Figure \ref{fig:inspectAI}) presents each INSPECT-SR assessment check together with the collected evidence and suggested outcomes. Human reviewers accept, modify, or override these suggestions before submitting their final assessment of the publication and the comment explaining their rationale. All provenance information related to evidence items, assessment data, and agents participating in the assessment process is logged in a PostgreSQL database.

\subsection{Pilot Deployment of INSPECT-AI }\label{sec:deployment}

The INSPECT-AI tool was piloted with 13 volunteers recruited from the Institute of Applied Health Sciences, University of Aberdeen, Centre for Biostatistics, University of Manchester, and the Indiana University School of Public Health-Bloomington during October to November 2025. \rev{The pilot study was designed to assess feasibility and usability of INSPECT-AI. All 13 participants completed a user feedback questionnaire which used standard response scales adapted to the question content (e.g., ‘Never’, ‘Occasionally’, ‘Frequently’, and ‘Very frequently’).} \rev{The participants had varied professional roles and were distinct from the biomedical researchers and research sleuths in the core team who also contributed additional assessments included in RIPE-KG.}  Participants used INSPECT-AI independently to assess 69 publications and generated 104 assessment traces (i.e., some publications were assessed by more than one reviewer). We provided the participants with a selection of publications with known trustworthiness issues based on former research; however, the participants were also encouraged to select their own publications. Most participants (77\%) had experience of conducting systematic reviews, with 31\% of participants reporting that they conduct reviews frequently (n=2) or very frequently (n=2), and (38.5\%) had ten or more years’ experience of conducting reviews. Only 61.5\% of participants had previously undertaken integrity assessments of RCTs, with more (77\%) having experience of conducting risk of bias assessments of randomised controlled trials (RCTs). Just under half of the participants (46\%) found the tool easy (n=3) or very easy to use (n=3) and 69\% found the tool either fairly user-friendly (n=6) or very user-friendly (n=3). None of the participants reported that they found the tool difficult to use or that the tool was not user-friendly. The response time of the tool was rated as quick (n=5) or very quick (n=4) by 69\% of participants.  
Although the tool could not completely remove the complexity of the assessment process, the tool's usability and speed of processing were accepted by most users. It was important to validate community acceptance of the INSPECT-AI tool as the information processed by the tool framed the development of RIPE-O (Section \ref{sec:ontology}) and the data collected during this study formed the basis of RIPE-KG (Section \ref{sec:kg}).

\section{Research Integrity Provenance \& Evidence Ontology (RIPE-O)} \label{sec:ontology}

\begin{table}[h]
\caption{\rev{Prefixes used in the remainder of this paper.}}
\label{tab:prefixes}
\centering
{\scriptsize

\begin{tabular}{ll}
\hline
\textbf{Prefix} & \textbf{Namespace} \\
\hline
ripe   & https://w3id.org/ripe/ripe-o\# \\
tido  & https://w3id.org/tido\# \\
prov  & http://www.w3.org/ns/prov\# \\
soa   & https://semopenalex.org/ontology/ \\
cito  & http://purl.org/spar/cito/ \\
fabio & http://purl.org/spar/fabio/ \\
dcterms & http://purl.org/dc/terms/ \\
foaf  & http://xmlns.com/foaf/0.1/ \\
prism & http://prismstandard.org/namespaces/basic/3.0/ \\
org   & http://www.w3.org/ns/org\# \\
xsd   & http://www.w3.org/2001/XMLSchema\# \\
schema & https://schema.org/ \\
rdfs  & http://www.w3.org/2000/01/rdf-schema\# \\
owl   & http://www.w3.org/2002/07/owl\# \\
ripekg &  https://w3id.org/ripe/ripe-kg/ \\
\hline
\end{tabular}
}
\end{table}

The aim of Research Integrity Provenance \& Evidence Ontology (RIPE-O) is to document provenance of the research integrity assessment process combining automated and human-generated outputs. Although the ontology design has been driven by the INSPECT-AI use case, the modelled concepts remain generic. \rev{RIPE-O does not prescribe which questions are investigated. Although its initial design was driven by INSPECT-SR, the same provenance pattern can document other investigations related to research integrity by defining new questions.} We also argue that various INSPECT-AI checks such as those concerning retraction notices, peer comments, and expressions of concern, are universally applicable to all academic literature.   

\begin{figure}[htbp]
\centering
\includegraphics[width=\textwidth]{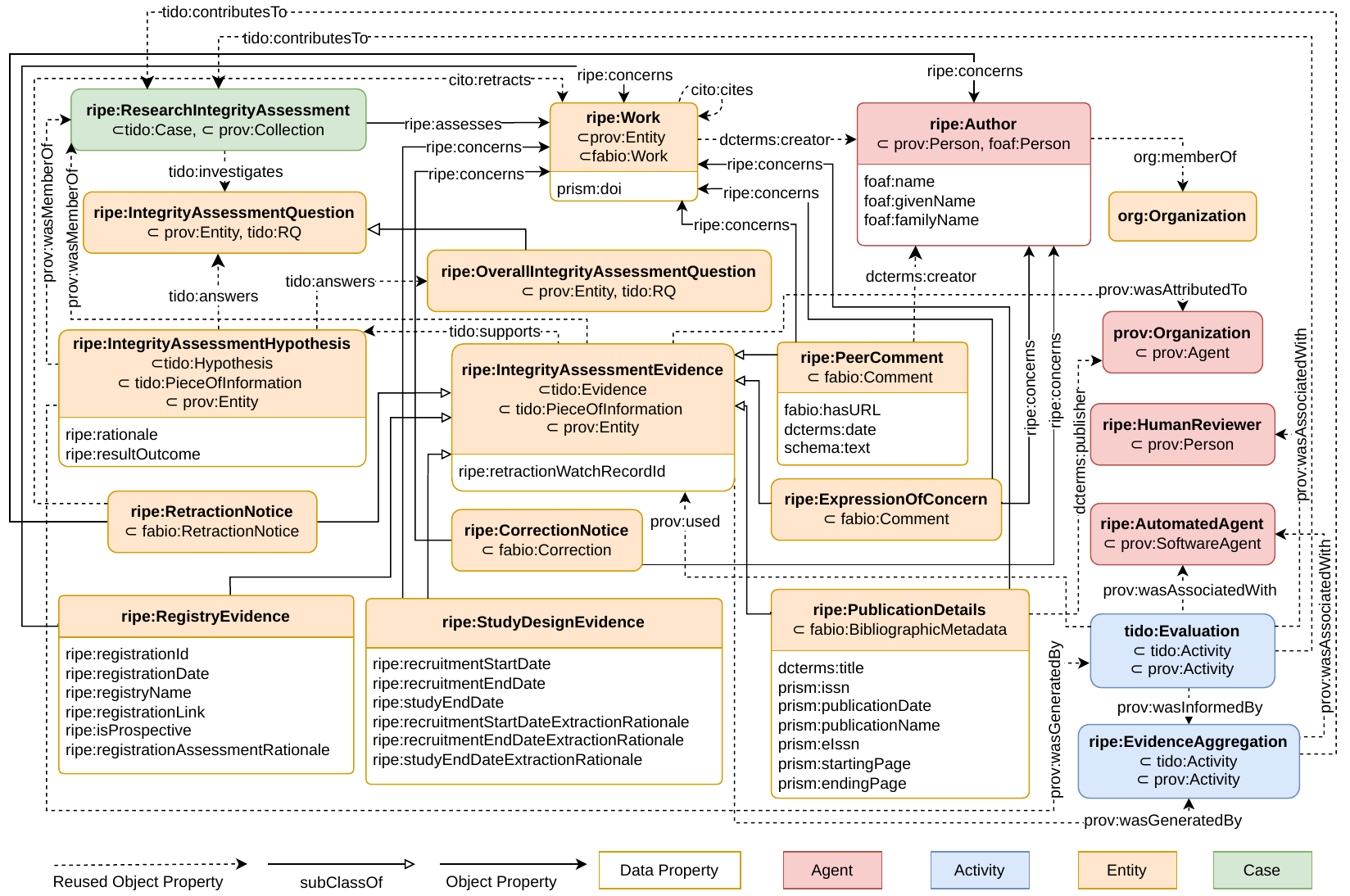}
\caption{The RIPE-O Ontology.}
\label{fig:ontology}
\end{figure}

Figure~\ref{fig:ontology} illustrates the core classes and relationships of the ontology. RIPE-O reuses concepts and properties from SPAR ontologies, namely FABIO and CITO~\cite{peroni2018spar} to model generic bibliographic data. Furthermore, RIPE-O extends TIDO \cite{roothaert2025tido}, which provides a vocabulary for modelling decision-making processes, including concepts such as evidence evaluation activities, evidence, investigated research questions, and hypotheses. TIDO itself extends  W3C PROV-O~\cite{W3C_PROV_O} and therefore the provenance traces are documented using activities, entities and agents. This allows us to describe retrospective causal graphs where activities use and generate entities, with agents bearing some responsibility for activities taking place and sharing attributions for the generated entities.

\begin{figure}[htbp]
\centering
\includegraphics[width=\textwidth]{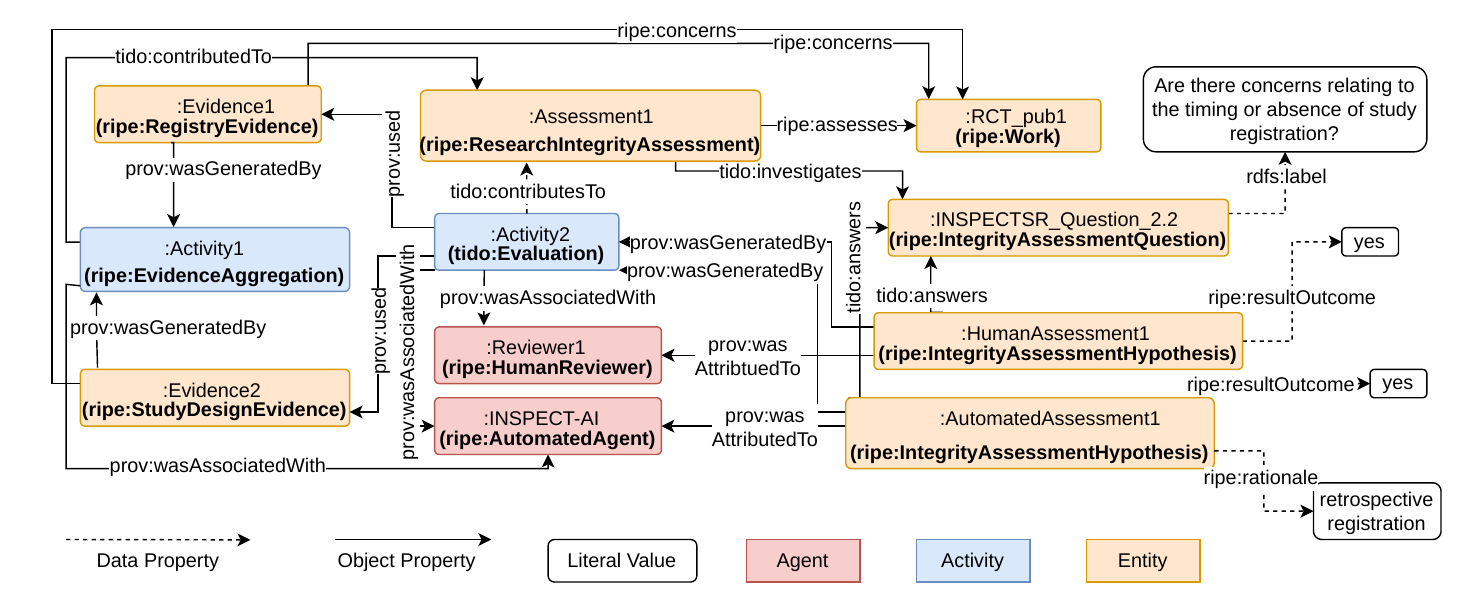}
\caption{An example partial provenance trace documenting considered evidence and assessment results for question about concerns related to the timing of the trial registration.}
\label{fig:ripe-kg assessment}
\end{figure}

To illustrate the application of RIPE-O (Figure \ref{fig:ripe-kg assessment}), consider an RCT publication (\texttt{ripe:Work}) that is assessed by a series of interdependent activities performed by human reviewers (\texttt{ripe:HumanReviewer}) and automated agents (\texttt{ripe:AutomatedAgent}) contributing to the assessment (\texttt{ripe:ResearchInte\allowbreak grityAssessment}). Each assessment investigates a number of questions (\texttt{ripe: IntegrityAssessmentQuestion}). For example, INSPECT-SR check 2.2 requires the assessor to determine whether the publication has any associated concerns relating to the timing or absence of study registration.
These questions are answered by hypotheses  (\texttt{ripe:Integrity\allowbreak AssessmentHypothesis}) which are results of the evaluation  activity (\texttt{tido:Eva\allowbreak luation}). 
The hypotheses are generated following the evaluation of  evidence (\texttt{ripe:Integrity\allowbreak AssessmentEvidence}) which is generated by the INSPECT-AI's automated  evidence aggregation activity (\texttt{ripe:EvidenceAggregation}).  
The evidence may include standard notices such as \texttt{ripe:CorrectionNotice},
\texttt{ripe:RetractionNotice}, or \texttt{ripe:Expre\-ssionOfConcern} but also evidence generated from the publication's content or external resources. This includes trial design details (\texttt{ripe:Study\allowbreak DesignEvidence}) such as start and end dates of participant recruitment mentioned in the article, and trial registration data (\texttt{ripe:RegistryEvidence}) extracted from the trial registries. The ontology  also models \texttt{ripe:PeerComment} which is a common post-publication review method allowing the community to comment on already published articles.    

We  used the OntOlogy Pitfall Scanner (OOPS!) \cite{poveda2014oops} to detect potential modelling errors and validated the ontology by converting the competency questions into SPARQL queries.\footnote{\url{https://github.com/RIPE-Observatory/RIPE-O/tree/main/cqs}} The results were then assessed against real-world data generated during the evaluation of the INSPECT-AI tool, which were transformed into RIPE-KG (discussed in the next section).

\section{Research Integrity Provenance and Evidence Knowledge Graph (RIPE-KG)} \label{sec:kg}

\begin{figure}[htbp]
      \centering
      \includegraphics[width=\textwidth]{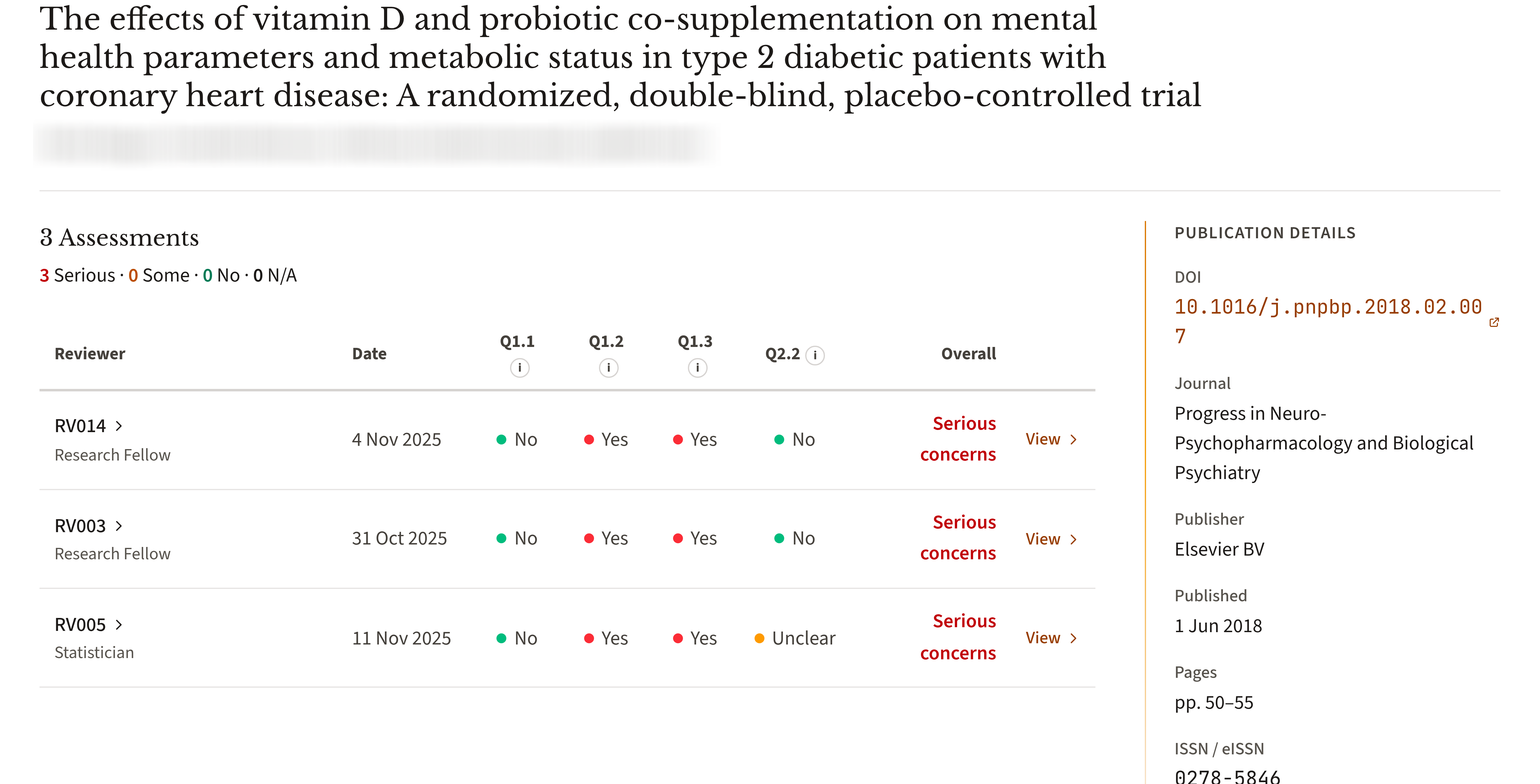}
      \caption{A partial view of the RIPE-KG web interface. Basic bibliographic information about the publication is displayed alongside three research integrity assessments. The view can be expanded to allow inspection of the individual pieces of evidence considered for each check.}
      \label{fig:KG-gui}
\end{figure}

RIPE-KG contains provenance traces of research integrity assessments performed using INSPECT-AI during tool evaluation (see Section \ref{sec:deployment}), as well as assessments performed by our expert team members (research sleuths). \rev{To protect reviewer anonymity in RIPE-KG while distinguishing assessments performed by different reviewers, each reviewer is identified by a pseudonymised reviewer ID and their role.} The KG can be accessed via SPARQL endpoint or via web-based GUI (Figure \ref{fig:KG-gui}). At the time of writing, RIPE-KG contains 140 assessments of 95 distinct publications, 927 distinct author entities representing the authors of  assessed publications, references and PubPeer comments, and  1{,}221 research integrity hypotheses (INSPECT-SR check outcomes) generated by both the tool and human reviewers.
To generate the KG, we transform INSPECT-AI's JSON logs to RDF using YARRRML mappings.~\footnote{\url{https://github.com/RIPE-Observatory/RIPE-KG/blob/main/mappings/ripe.yarrrml.yml}} 

Author identities are disambiguated during graph construction to avoid potential duplication due to inconsistent author names across different publications and authors sharing the same name. INSPECT-AI uses GROBID to extract the author list and affiliations from the publication PDF which are assigned a local IRI. Where DOI of the publication is known, we use the OpenAlex API~\cite{openalexapi} to retrieve OpenAlex identifiers for the authors.~\footnote{We have found OpenAlex API to be more reliable than the SemOpenAlex SPARQL endpoint which at times provided incomplete results.} OpenAlex identifiers are then converted to their corresponding SemOpenAlex author IRIs which are linked using \texttt{owl:sameAs} to the local author instance, so that the published graph remains aligned with SemOpenAlex and can support federated queries. The local author instance is linked to the SemOpenAlex author IRI only when the author position, surname, and given name match, with the exception of cases where initials are matched to the first name, where  the author position, surname, and affiliation also match.
For the current RIPE-KG, GROBID extracted 1{,}202 publication-author mentions from the assessed publications, which were disambiguated into 875 distinct RIPE author instances. Of these, 766 are linked to SemOpenAlex, \rev{while 109 (12.5\%) remain unlinked due to missing OpenAlex author metadata for the relevant DOI or because the OpenAlex author list did not match the metadata extracted from the publication.} 

\subsection{Analysing Assessment Outputs}

 INSPECT-AI currently implements four INSPECT-SR checks recorded as research integrity questions in each of the assessments. These are: \textit{Does the study have an associated retraction?} \textit{Does the study have an associated expression of concern or other relevant post-publication notice?} \textit{Do other studies by the research team highlight causes for concern (associated retractions, expressions of concern, relevant post-publication notices)?}  \textit{Are there concerns relating to the timing or absence of study registration?} The results of individual investigated questions can be easily retrieved from KG. Listing \ref{lst:human-vs-automated} shows an example SPARQL query that can be used to  compare the results of automated and human-reviewed hypotheses for the same integrity question within individual assessments. \rev{The query links both hypotheses to the same question and assessment through \texttt{tido:answers} and \texttt{prov:wasMemberOf}, and distinguishes them through \texttt{prov:wasAttributedTo} links to an automated agent or human reviewer.} Differences in these values identify cases where the human reviewer disagreed with the automated outcome and provided justification.

\medskip
{\scriptsize
\begin{lstlisting}[caption={SPARQL query identifying \texttt{ripe:ResearchIntegrityAssessment} instances where human-reviewed and automated outcomes differ and a human rationale is recorded. Example results shown in Table \ref{tab:automated-human-disagreements}.}, label={lst:human-vs-automated}]
SELECT DISTINCT ?assessment ?questionText ?automatedOutcome ?humanOutcome
       ?humanRationale
WHERE {
  ?assessment a ripe:ResearchIntegrityAssessment.
  ?question a ripe:IntegrityAssessmentQuestion;
      rdfs:label ?questionText.
  ?automatedHypothesis tido:answers ?question;
      ripe:resultOutcome ?automatedOutcome;
      prov:wasMemberOf ?assessment;
      prov:wasAttributedTo ?automatedAgent.
  ?automatedAgent a ripe:AutomatedAgent.
  ?humanHypothesis tido:answers ?question;
      ripe:resultOutcome ?humanOutcome;
      ripe:rationale ?humanRationale;
      prov:wasMemberOf ?assessment;
      prov:wasAttributedTo ?humanReviewer.
  ?humanReviewer a ripe:HumanReviewer.
  FILTER(?automatedOutcome != ?humanOutcome && STR(?humanRationale) != "")
}
ORDER BY ?assessment ?questionText

\end{lstlisting}
}

\begin{table}[]
\caption{\rev{Example RIPE-KG results listing disagreements between automated and human outcomes for integrity assessment questions.}}
\label{tab:automated-human-disagreements}
\centering
\scriptsize
\setlength{\tabcolsep}{2pt}
\begin{tabularx}{\linewidth}{|p{0.25\linewidth}|X|p{0.12\linewidth}|p{0.10\linewidth}|X|}
\toprule
\textbf{Assessment} & \textbf{Question} & \parbox[t]{\linewidth}{\centering\textbf{Automated}\\\textbf{Outcome}} & \parbox[t]{\linewidth}{\centering\textbf{Human}\\\textbf{Outcome}} & \textbf{Human Rationale} \\
\midrule
\begin{tabular}[t]{@{}l@{}}RIPEA0E351EBF25A8\\D2C5\end{tabular} & 1.3. Do other studies by the research team highlight causes for concern? & yes & no & It looks like a different Fatemeh Ghasemi who is flagged in retraction watch (the retraction watch Fatemeh is a computer engineer). \\
\midrule

\begin{tabular}[t]{@{}l@{}}RIPEAF6F62EBFADD4\\2DFB\end{tabular} & 1.3. Do other studies by the research team highlight causes for concern? & yes & no & Although a correction has been flagged, this relates to the failure to acknowledge a particular individual and is not related to the integrity of the study results. \\
\bottomrule
\end{tabularx}
\end{table}

We have analysed 514 assessments~$\times$~question pairs\footnote{The reviewers are not required to answer all assessment questions if significant issues (e.g., retraction notice) are found early on.} for which both automated and human-reviewed outcomes are available. Of these, 444 (86.4\%) show agreement between automated and human-reviewed outcomes, while 70 (13.6\%) show disagreement. The disagreement pattern is uneven across questions: retraction status shows 11 disagreements out of 129 pairs (8.5\%), post-publication notices 5 out of 127 (3.9\%), research-team-related concerns 20 out of 128 (15.6\%), and study registration 34 out of 130 (26.2\%). 

However, human reviews also do not appear to agree consistently. We have analysed 22 publications where more than one assessment was recorded. In the context of individual investigated questions for each of these publications, reviewers agree in 15 of 18 cases on retraction status (83.3\%), 16 of 19 on post-publication notices (84.2\%), 17 of 18 on research-team-related concerns (94.4\%), but only 8 of 18 on study registration concerns (44.4\%). These results suggest that study registration is the least agreeable question because, unlike retractions or post-publication notices, it often requires expert domain knowledge to determine whether the reported registration, recruitment, and study timelines are compatible with acceptable clinical trial practice. Disagreements on the retraction status are also interesting, because this may indicate that some human reviewers sided with the automated INSPECT-AI suggestions without following the additional guidance to check publishers' websites. This extra step is suggested due to the current INSPECT-AI inability to bypass the firewall protections of various publishers' websites and, hence, can retrieve the retraction status only from the PDF of the publication or the Retraction Watch Database, which may be incomplete~\cite{Bakker121}. The tool also highlights retracted references and does not  automatically suggest concerns; however, leaves the decision to the human reviewer.

The outcomes of the four individual integrity questions are considered before answering the overall assessment question: \textit{What is the overall integrity assessment of the study?}
INSPECT-AI currently generates automated overall hypotheses only if retractions of the main publication have been detected and thus most of the overall assessments in our KG are solely human-generated.
RIPE-KG contains 140 human-reviewed overall assessments for 95 works. Of these, 22 works were reviewed more than once. Among those 22 repeatedly reviewed works, 16 (72.7\%) received the same overall human assessment across repeated reviews, while 6 (27.3\%) showed disagreement between reviewers. The clearest disagreement example is DOI \texttt{10.1111/bjdp.12503} \footnote{\url{https://ripe-kg.inspectai.app/publications/10.1111\%2Fbjdp.12503}}, which appears in 8 human-reviewed assessments spanning all three overall assessment categories: \texttt{no-concerns}, \texttt{some-concerns}, and \texttt{serious-concerns}. 

Overall, the results support the need for provenance tracking in the research integrity assessment context due to disagreements not only among human assessors but also between human and automated assessments. While more advanced techniques for assessing the trustworthiness of assessment results based on detailed provenance traces were beyond the scope of this paper, they represent a potential direction for future work.

\subsection{Author Networks}

The RIPE-KG also supports basic author-network analysis. For example, it is possible to identify authors who frequently co-author publications with integrity concerns. Listing \ref{lst:author-network} shows an example SPARQL query that combines the overall human integrity outcome with authorship links to count distinct publications shared by each author pair. 

\medskip
{\scriptsize
\begin{lstlisting}[caption={SPARQL query identifying author pairs that frequently co-author publications which have received at least one overall assessment highlighting \texttt{serious-concerns}. }, label={lst:author-network}]
SELECT ?authorA ?authorB (COUNT(DISTINCT ?work) AS ?sharedWorks)
WHERE {
  ?assessment a ripe:ResearchIntegrityAssessment;
      ripe:assesses ?work .
  ?overallQuestion a ripe:OverallIntegrityAssessmentQuestion .
  ?hypothesis tido:answers ?overallQuestion;
      ripe:resultOutcome "serious-concerns";
      prov:wasMemberOf ?assessment;
      prov:wasAttributedTo ?humanReviewer.
  ?humanReviewer a ripe:HumanReviewer.
  ?work dcterms:creator ?authorAResource, ?authorBResource .
  ?authorAResource foaf:name ?authorA.
  ?authorBResource foaf:name ?authorB.
  FILTER(?authorA < ?authorB)
}
GROUP BY ?authorAResource ?authorBResource ?authorA ?authorB
ORDER BY DESC(?sharedWorks) ?authorA ?authorB
\end{lstlisting}
}


\subsection{Integrating External Scholarly Knowledge Graphs}

RIPE-KG can be integrated with data from other scholarly knowledge graphs to answer questions such as ``What are the research integrity assessment results of publications from a specific field of study?'' Listing \ref{lst:field-integrity-query} shows a federated query using the SemOpenAlex knowledge graph to retrieve field classification information for assessed publications. \rev{The \texttt{owl:sameAs} link identifies the corresponding SemOpenAlex work, while the \texttt{SERVICE} clause retrieves its field classification and the remainder of the query retrieves the human-reviewed overall outcome from RIPE-KG.}

{\scriptsize
\begin{lstlisting}[caption={Federated SPARQL query linking RIPE works with SemOpenAlex classification concept	
C134018914 (Endocrinology) and retrieving human-reviewed overall outcomes. Example results shown in Table \ref{tab:field-integrity-endocrinology}.}, label={lst:field-integrity-query}]

SELECT DISTINCT ?work ?doi ?title ?overallOutcome
WHERE {
  ?assessment a ripe:ResearchIntegrityAssessment;
      ripe:assesses ?work .
  ?work owl:sameAs ?soaWork .
  OPTIONAL { ?work prism:doi ?doi }
  OPTIONAL { ?work dcterms:title ?title }

  SERVICE <https://semopenalex.org/sparql> {
    ?soaWork soa:hasConcept <https://semopenalex.org/concept/C134018914>.
  }

  ?overallQuestion a ripe:OverallIntegrityAssessmentQuestion .
  ?overallHypothesis tido:answers ?overallQuestion;
      ripe:resultOutcome ?overallOutcome;
      prov:wasMemberOf ?assessment;
      prov:wasAttributedTo ?humanReviewer .
  ?humanReviewer a ripe:HumanReviewer .
}
ORDER BY ?doi ?overallOutcome
\end{lstlisting}
}


\begin{table}[]
\caption{\rev{Example RIPE-KG results listing  endocrinology works with human-reviewed overall integrity outcomes.}}
\label{tab:field-integrity-endocrinology}
\centering
\scriptsize
\begin{tabularx}{\linewidth}{X |X| c}
\toprule
\textbf{DOI} & \textbf{Title} & \textbf{Overall Outcome} \\
\midrule
10.1002/ptr.6406 & Hesperidin improves hepatic steatosis, hepatic enzymes, and metabolic and inflammatory parameters in patients with nonalcoholic fatty liver disease: A randomized, placebo-controlled, double-blind clinical trial & some-concerns \\
\midrule
10.1007/s00394-018-1760-8 & Effects of vitamin D supplementation on metabolic and endocrine parameters in PCOS: a randomized-controlled trial & some-concerns \\
\bottomrule
\end{tabularx}
\end{table}

\subsection{Costs}
During the pilot study period (October-November 2025), the total operational cost of INSPECT-AI was approximately US\$79.53. The largest component was infrastructure: all services ran as Docker containers on a single VPS (8 vCPU, 32 GB RAM), costing \$77 for the period. LLM costs were modest with Gemini 2.0 Flash, accessed through Google AI Studio, handled structured data extraction from PDFs at a cost of \$1.34. At this level of expenditure, the marginal cost per paper analysed was below \$0.10.

\section{Limitations}

The current version of the INSPECT-AI prototype does not provide AI support for all INSPECT-SR checks. In addition, because of our reliance on OpenAlex for author disambiguation, RIPE-KG will inherit any insufficiencies present in this external database. The functionality of the tool is also affected by the quality of external information sources such as SemOpenAlex knowledge graph, Retraction Watch Database, and PubPeer.

\section{Conclusions}

In this paper, we introduced the first end-to-end pipeline for generating provenance traces of research integrity assessments and publishing them as open knowledge graphs in accordance with FAIR principles. \rev{Our approach advances the wider agenda of scholarly knowledge graphs and provides reusable technologies for integrating outputs from existing and future research integrity assessment tools into provenance knowledge graphs that can support new research integrity metrics and research-output appraisal by funders, publishers, and other end users.}

Our findings highlight the inherent complexity and variability of research integrity assessment, with discrepancies between human and automated assessments. These results support the need to incorporate human validation into LLM-based decision support systems to ensure the reliability and accountability of their outputs. Given the limited availability of human experts in this context, the proposed approach is particularly relevant, as it enables the preservation and sharing of validated assessments. As the KG grows over time through the addition of tools covering more disciplines, we envision this resource becoming a key enabler of new semantic approaches for transparent research integrity assessments. 

\section{Future Work}

In future work, we will focus on extending the coverage of INSPECT-SR checks within the INSPECT-AI tool, as well as exploring the application of RIPE-O for documenting research integrity assessments in a broader biomedical context and across other disciplines. We will continue to engage with systematic reviewers and research sleuths through additional community-based deployments of INSPECT-AI to expand the RIPE-KG knowledge graph. \rev{The growth of RIPE-KG will also be supported through continuous assessments performed by members of our core expert team.} By targeting specific trusted user groups, we will maintain the quality and relevance of the RIPE-KG and reduce the potential for malicious content. We will also investigate advanced graph-based  selection mechanisms that leverage RIPE-KG data to automatically discover potentially untrustworthy publications for future human-based assessments.

\section{Acknowledgments}
This work was supported by the Economic and Social Research Council (ESRC)
[grant number UKRI1083]. We also acknowledge the generous support of alumni and friends in establishing the University of Aberdeen’s Interdisciplinary Institute, which partly enabled this research, including Dr Jane Hellman Caseley (MBChB 1956), Professor Patrick Meares (DSc 1959), Nancy Miller (MA 1942), Norman Robertson, Dr Ian Slessor (MBChB 1956) and Anne Young (MA 1957). For the purpose of open access, the author has applied a Creative Commons Attribution (CC BY) licence to any Author Accepted Manuscript version arising from this submission.

\section{Declaration of use of Generative AI}
GenAI tools have been used to generate LaTeX formatting code for tables, figures and listings and for text shortening suggestions. GenAI tools have also been used to support programming activities including query and code generation during the development of INSPECT-AI and RIPE-KG web interface. 
%
%
%
 \bibliographystyle{splncs04}
 \bibliography{bibliography_rebuttal_revision}

@InProceedings{nielsen2017scholia,
doi = "10.1007/978-3-319-70407-4_36",
author="Nielsen, Finn {\AA}rup
and Mietchen, Daniel
and Willighagen, Egon",
editor="Blomqvist, Eva
and Hose, Katja
and Paulheim, Heiko
and {\L}awrynowicz, Agnieszka
and Ciravegna, Fabio
and Hartig, Olaf",
title="Scholia, Scientometrics and Wikidata",
booktitle="The Semantic Web: ESWC 2017 Satellite Events",
year="2017",
publisher="Springer",
address="Cham",
pages="237--259",
isbn="978-3-319-70407-4"
}

@techreport{W3C_PROV_O,
  author      = {Lebo, Timothy and Sahoo, Satya and McGuinness, Deborah},
  title       = {{PROV-O}: The {PROV} Ontology},
  url         = {http://www.w3.org/TR/2013/REC-prov-o-20130430/},
  type        = {{W3C} Recommendation},
  institution = {W3C},
  year        = {2013},
  month       = {Apr},
  day         = {30}
}

@article {Bakker121,
	author = {Bakker, Caitlin and Boughton, Stephanie and Faggion, Clovis Mariano and Fanelli, Daniele and Kaiser, Kathryn and Schneider, Jodi},
	title = {Reducing the residue of retractions in evidence synthesis: ways to minimise inappropriate citation and use of retracted data},
	volume = {29},
	number = {2}, 
	pages = {121--126},
	year = {2024},
	doi = {10.1136/bmjebm-2022-111921},
	publisher = {Royal Society of Medicine},
	issn = {2515-446X},
	URL = {https://ebm.bmj.com/content/29/2/121},
	eprint = {https://ebm.bmj.com/content/29/2/121.full.pdf},
	journal = {BMJ Evidence-Based Medicine}
}

@Article{poveda2014oops,
author={Poveda-Villal{\'o}n, Mar{\'i}a
and G{\'o}mez-P{\'e}rez, Asunci{\'o}n
and Su{\'a}rez-Figueroa, Mari Carmen},
title={OOPS! (OntOlogy Pitfall Scanner!): An On-line Tool for Ontology Evaluation},
journal={International Journal on Semantic Web and Information Systems (IJSWIS)},
year={2014},
publisher={IGI Global Scientific Publishing},
address={Hershey, PA, USA},
volume={10},
number={2},
pages={7-34},
issn={1552-6283},
doi={10.4018/ijswis.2014040102}
}

@inproceedings{10.5555/2889875.2889882,
author = {Ram, Sudha and Liu, Jun},
title = {A new perspective on semantics of data provenance},
year = {2009},
publisher = {CEUR-WS.org},
address = {Aachen, DEU},
booktitle = {Proceedings of the First International Conference on Semantic Web in Provenance Management - Volume 526},
pages = {35–40},
numpages = {6},
location = {Washington DC},
series = {SWPM'09}
}

@Article{verma2023scholarly,
author={Verma, Shilpa
and Bhatia, Rajesh
and Harit, Sandeep
and Batish, Sanjay},
title={Scholarly knowledge graphs through structuring scholarly communication: a review},
journal={Complex {\&} Intelligent Systems},
year={2023},
month={Feb},
day={01},
volume={9},
number={1},
pages={1059-1095},
issn={2198-6053},
doi={10.1007/s40747-022-00806-6}
}

@article{protege,
  author    = {Mark A. Musen},
  title     = {The prot{\'{e}}g{\'{e}} project: a look back and a look
               forward},
  journal   = {{AI} Matters},
  volume    = {1},
  number    = {4},
  pages     = {4--12},
  year      = {2015},
  doi       = {10.1145/2757001.2757003},
  bibsource = {dblp computer science bibliography, https://dblp.org}
}

@inproceedings{garijo2017widoco,
  title={WIDOCO: a wizard for documenting ontologies},
  author={Garijo, Daniel},
  booktitle={International Semantic Web Conference},
  pages={94--102},
  year={2017},
  organization={Springer, Cham},
  doi = {10.1007/978-3-319-68204-4_9},
  funding = {USNSF ICER-1541029, NIH 1R01GM117097-01},
  url={http://dgarijo.com/papers/widoco-iswc2017.pdf}
}

@InProceedings{heyvaert2018declarative,
author="Heyvaert, Pieter
and De Meester, Ben
and Dimou, Anastasia
and Verborgh, Ruben",
editor="Gangemi, Aldo
and Gentile, Anna Lisa
and Nuzzolese, Andrea Giovanni
and Rudolph, Sebastian
and Maleshkova, Maria
and Paulheim, Heiko
and Pan, Jeff Z.
and Alam, Mehwish",
title="Declarative Rules for Linked Data Generation at Your Fingertips!",
booktitle="European Semantic Web Conference",
year="2018",
publisher="Springer",
address="Cham",
pages="213--217",
isbn="978-3-319-98192-5",
doi="10.1007/978-3-319-98192-5_40"
}

@inproceedings{peroni2018spar,
author = {Peroni, Silvio and Shotton, David},
title = {The SPAR Ontologies},
year = {2018},
isbn = {978-3-030-00667-9},
publisher = {Springer-Verlag},
address = {Berlin, Heidelberg},
doi = {10.1007/978-3-030-00668-6_8},
booktitle = {The Semantic Web – ISWC 2018: 17th International Semantic Web Conference, Monterey, CA, USA, October 8–12, 2018, Proceedings, Part II},
pages = {119–136},
numpages = {18},
location = {Monterey, CA, USA}
}

@misc{cochrane_methods,
  author       = {{Cochrane}},
  title        = {Methods in Cochrane},
  howpublished = {\url{https://www.cochrane.org/authors/methods-cochrane}},
  year         = {n.d.},
  note         = {Accessed: 2026-04-27}
}

@misc{UKRIO2025,
  author    = {{UK Research Integrity Office (UKRIO)}},
  title     = {Introductory guide to the Concordat to Support Research Integrity},
  year      = {2025},
  url       = {https://ukrio.org/research-integrity/the-concordat-to-support-research-integrity/introductory-guide-to-the-concordat-to-support-research-integrity/},
  note      = {Accessed: 14 April 2026}
}

@inproceedings{roothaert2025tido,
author = {Roothaert, Ritten and Schlobach, Stefan and Massacci, Fabio and Stork, Lise},
title = {TIDO: The Threat Intelligence Decision Ontology},
year = {2025},
isbn = {9798400718670},
publisher = {Association for Computing Machinery},
address = {New York, NY, USA},
doi = {10.1145/3731443.3771351},
booktitle = {Proceedings of the 13th Knowledge Capture Conference 2025},
pages = {82–89},
numpages = {8},
series = {K-CAP '25}
}

@article {Wilkinson2025.09.03.25334905,
	author = {Wilkinson, Jack and Heal, Calvin and Flemyng, Ella and Antoniou, Georgios A. and Aburrow, Tony and Alfirevic, Zarko and Avenell, Alison and Barbour, Virginia and Berghella, Vincenzo and Bishop, Dorothy V. M. and Bordewijk, Esm{\'e}e M and Brown, Nicholas J. L. and Christopher, Jana and Clarke, Mike and Dahly, Darren and Dennis, Jane and Dicker, Patrick and Dumville, Jo and Frankish, Helen and Grey, Andrew and Grohmann, Steph and Gurrin, Lyle C. and Hayden, Jill A. and Heathers, James A.J. and Hunter, Kylie E and Hussey, Ian and Jung, Lukas and Lam, Emily and Lasserson, Toby J. and Lensen, Sarah and Li, Tianjing and Li, Wentao and Liu, Jianping and Loder, Elizabeth and Lundh, Andreas and Meyerowitz-Katz, Gideon and Mol, Ben W. and Naudet, Florian and Noel-Storr, Anna and O{\textquoteright}Connell, Neil E. and Parker, Lisa and Redberg, Rita F. and Redman, Barbara K. and Richardson, Rachel and Seidler, Anna Lene and Sheldrick, Kyle and Sydenham, Emma and Wely, Madelon van and Vorland, Colby J. and Wang, Rui and Weibel, Stephanie and Wjst, Matthias and Bero, Lisa and Kirkham, Jamie J.},
	title = {INSPECT-SR: a tool for assessing trustworthiness of randomised controlled trials},
	elocation-id = {2025.09.03.25334905},
	year = {2025},
	doi = {10.1101/2025.09.03.25334905},
	publisher = {Cold Spring Harbor Laboratory Press},
	URL = {https://www.medrxiv.org/content/early/2025/10/21/2025.09.03.25334905},
	eprint = {https://www.medrxiv.org/content/early/2025/10/21/2025.09.03.25334905.full.pdf},
	journal = {medRxiv}
}

@article{POVEDAVILLALON2022104755,
  title    = {LOT: An industrial oriented ontology engineering framework},
  journal  = {Engineering Applications of Artificial Intelligence},
  volume   = {111},
  pages    = {104755},
  year     = {2022},
  issn     = {0952-1976},
  doi      = {10.1016/j.engappai.2022.104755},
  author   = {María Poveda-Villalón and Alba Fernández-Izquierdo and Mariano Fernández-López and Raúl García-Castro}
}

@misc{manghi2019openaire,
  author       = {Manghi, Paolo and
                  Bardi, Alessia and
                  Atzori, Claudio and
                  Baglioni, Miriam and
                  Manola, Natalia and
                  Schirrwagen, Jochen and
                  Principe,  Pedro},
  title        = {The OpenAIRE Research Graph Data Model},
  month        = apr,
  year         = 2019,
  publisher    = {Zenodo},
  version      = {1.3},
  doi          = {10.5281/zenodo.2643199},
}

@article{grey2025inconsistency,
  title={Inconsistency in publishers' responses to integrity concerns about published research. Evidence and suggested improvements},
  author={Grey, Andrew and Avenell, Alison and Gaby, Alan and Bolland, Mark J},
  journal={Journal of Clinical Epidemiology},
  volume={186},
  year={2025},
  publisher={Elsevier},
  doi={10.1016/j.jclinepi.2025.111918}
}

@article{grey2025ten,
author={Grey, Andrew and Avenell, Alison and Bolland, Mark J},
title = {Ten Years later: Assessments of the integrity of publications from one research group with multiple retractions},
journal = {Accountability in Research},
volume = {32},
number = {4},
pages = {488--508},
year = {2025},
publisher = {Taylor \& Francis},
doi = {10.1080/08989621.2023.2295996}
}

@misc{retractionwatch2018sleuths,
  author       = {{Retraction Watch}},
  title        = {Meet the scientific sleuths: Ten who've had an impact on the scientific literature},
  year         = {2018},
  month        = jun,
  day          = {17},
  url          = {https://retractionwatch.com/2018/06/17/meet-the-scientific-sleuths-ten-whove-had-an-impact-on-the-scientific-literature/},
  note         = {Accessed: 2026-03-03}
}

@Article{wilkinson2016fair,
author={Wilkinson, Mark D.
and Dumontier, Michel
and Aalbersberg, IJsbrand Jan
and Appleton, Gabrielle
and Axton, Myles
and Baak, Arie
and Blomberg, Niklas
and Boiten, Jan-Willem
and da Silva Santos, Luiz Bonino
and Bourne, Philip E.
and Bouwman, Jildau
and Brookes, Anthony J.
and Clark, Tim
and Crosas, Merc{\`e}
and Dillo, Ingrid
and Dumon, Olivier
and Edmunds, Scott
and Evelo, Chris T.
and Finkers, Richard
and Gonzalez-Beltran, Alejandra
and Gray, Alasdair J.G.
and Groth, Paul
and Goble, Carole
and Grethe, Jeffrey S.
and Heringa, Jaap
and 't Hoen, Peter A.C
and Hooft, Rob
and Kuhn, Tobias
and Kok, Ruben
and Kok, Joost
and Lusher, Scott J.
and Martone, Maryann E.
and Mons, Albert
and Packer, Abel L.
and Persson, Bengt
and Rocca-Serra, Philippe
and Roos, Marco
and van Schaik, Rene
and Sansone, Susanna-Assunta
and Schultes, Erik
and Sengstag, Thierry
and Slater, Ted
and Strawn, George
and Swertz, Morris A.
and Thompson, Mark
and van der Lei, Johan
and van Mulligen, Erik
and Velterop, Jan
and Waagmeester, Andra
and Wittenburg, Peter
and Wolstencroft, Katherine
and Zhao, Jun
and Mons, Barend},
title={The FAIR Guiding Principles for scientific data management and stewardship},
journal={Scientific Data},
year={2016},
month={Mar},
day={15},
volume={3},
number={1},
pages={160018},
issn={2052-4463},
doi={10.1038/sdata.2016.18}
}

@article{weeks2023trustworthiness,
author = {Weeks, Jo and Cuthbert, Anna and Alfirevic, Zarko},
title = {Trustworthiness assessment as an inclusion criterion for systematic reviews—What is the impact on results?},
journal = {Cochrane Evidence Synthesis and Methods},
volume = {1},
number = {10},
pages = {e12037},
doi = {10.1002/cesm.12037},
year = {2023},
publisher={Wiley Online Library}
}

@article{mol2023tract,
  title={Checklist to assess Trustworthiness in RAndomised Controlled Trials ({TRACT} checklist): concept proposal and pilot},
  author={Mol, Ben W. and Lai, Shimona and Rahim, Ayesha and Bordewijk, Esm{\'e}e M. and Wang, Rui and van Eekelen, Rik and Gurrin, Lyle C. and Thornton, Jim G. and van Wely, Madelon and Li, Wentao},
  journal={Research Integrity and Peer Review},
  volume={8},
  number={1},
  pages={6},
  year={2023},
  doi={10.1186/s41073-023-00130-8}
}

@article{weibel2023ria,
  title={Identifying and managing problematic trials: A research integrity assessment tool for randomized controlled trials in evidence synthesis},
  author={Weibel, Stephanie and Popp, Maria and Reis, Stefanie and Skoetz, Nicole and Garner, Paul and Sydenham, Emma},
  journal={Research Synthesis Methods},
  volume={14},
  number={3},
  pages={357--369},
  year={2023},
  doi={10.1002/jrsm.1599}
}

@article{grey2020reappraised,
  title={Check for publication integrity before misconduct},
  author={Grey, Andrew and Bolland, Mark J. and Avenell, Alison and Klein, Andrew A. and Gunsalus, C. K.},
  journal={Nature},
  volume={577},
  number={7789},
  pages={167--169},
  year={2020},
  doi={10.1038/d41586-019-03959-6}
}

@article{au2025tract,
  title={Using artificial intelligence to semi-automate trustworthiness assessment of randomized controlled trials: a case study},
  author={Au, Ling Shan and Qu, Lizhen and Nielsen, Jeremy and Ge, Zongyuan and Gurrin, Lyle C. and Mol, Ben W. and Wang, Rui},
  journal={Journal of Clinical Epidemiology},
  volume={180},
  pages={111672},
  year={2025},
  doi={10.1016/j.jclinepi.2025.111672}
}

@article{avenell2024randomized,
author={Avenell, Alison and Bolland, Mark J and Gamble, Greg D and Grey, Andrew},
title = {A randomized trial alerting authors, with or without coauthors or editors, that research they cited in systematic reviews and guidelines has been retracted},
journal = {Accountability in Research},
volume = {31},
number = {1},
pages = {14--37},
year = {2024},
publisher = {Taylor \& Francis},
doi = {10.1080/08989621.2022.2082290},

}

@article{kinney2023semantic,
  title={The Semantic Scholar Open Data Platform},
  author={Rodney Michael Kinney and Chloe Anastasiades and Russell Authur and Iz Beltagy and Jonathan Bragg and Alexandra Buraczynski and Isabel Cachola and Stefan Candra and Yoganand Chandrasekhar and Arman Cohan and Miles Crawford and Doug Downey and Jason Dunkelberger and Oren Etzioni and Rob Evans and Sergey Feldman and Joseph Gorney and David W. Graham and F.Q. Hu and Regan Huff and Daniel King and Sebastian Kohlmeier and Bailey Kuehl and Michael Langan and Daniel Lin and Haokun Liu and Kyle Lo and Jaron Lochner and Kelsey MacMillan and Tyler C. Murray and Christopher Newell and Smita Rao and Shaurya Rohatgi and Paul Sayre and Shannon Zejiang Shen and Amanpreet Singh and Luca Soldaini and Shivashankar Subramanian and A. Tanaka and Alex D Wade and Linda M. Wagner and Lucy Lu Wang and Christopher Wilhelm and Caroline Wu and Jiangjiang Yang and Angele Zamarron and Madeleine van Zuylen and Daniel S. Weld},
  journal={ArXiv},
  year={2023},
  volume={abs/2301.10140},
  url={https://api.semanticscholar.org/CorpusID:256194545}
}

@InProceedings{farber2023semopenalex,
author="F{\"a}rber, Michael
and Lamprecht, David
and Krause, Johan
and Aung, Linn
and Haase, Peter",
doi="10.1007/978-3-031-47243-5_6",
title="SemOpenAlex: The Scientific Landscape in 26 Billion RDF Triples",
booktitle="The Semantic Web -- ISWC 2023",
year="2023",
publisher="Springer",
address="Cham",
pages="94--112"
}

@inproceedings{jaradeh2019open,
author = {Jaradeh, Mohamad Yaser and Oelen, Allard and Farfar, Kheir Eddine and Prinz, Manuel and D'Souza, Jennifer and Kismih{\'o}k, G{\'a}bor and Stocker, Markus and Auer, S{\"o}ren},
title = {Open Research Knowledge Graph: Next Generation Infrastructure for Semantic Scholarly Knowledge},
year = {2019},
isbn = {9781450370080},
publisher = {Association for Computing Machinery},
address = {New York, NY, USA},
doi = {10.1145/3360901.3364435},
booktitle = {Proceedings of the 10th International Conference on Knowledge Capture},
pages = {243–246},
numpages = {4},
location = {Marina Del Rey, CA, USA},
series = {K-CAP '19}
}

@Article{peterson2023metascience,
author={Peterson, David
and Panofsky, Aaron},
title={Metascience as a Scientific Social Movement},
journal={Minerva},
year={2023},
month={Jun},
day={01},
volume={61},
number={2},
pages={147-174},
issn={1573-1871},
doi={10.1007/s11024-023-09490-3},
}

@InProceedings{grobid,
author="Lopez, Patrice",
editor="Agosti, Maristella
and Borbinha, Jos{\'e}
and Kapidakis, Sarantos
and Papatheodorou, Christos
and Tsakonas, Giannis",
title="GROBID: Combining Automatic Bibliographic Data Recognition and Term Extraction for Scholarship Publications",
booktitle="Research and Advanced Technology for Digital Libraries",
year="2009",
publisher="Springer Berlin Heidelberg",
address="Berlin, Heidelberg",
pages="473--474",
isbn="978-3-642-04346-8",
doi="10.1007/978-3-642-04346-8_62"
}

@misc{retractionWatchDatabase,
  title = {{Retraction Watch Database}},
  url   = {https://retractiondatabase.org/RetractionSearch.aspx?},
  key = {Retraction Watch Database},
  note  = {Last accessed 2026/04/27}
}

@misc{pubpeer,
  title = {{PubPeer}},
  url   = {https://pubpeer.com/},
  key = {PubPeer},
  note  = {Last accessed 2026/04/27}
}

@misc{openalexapi,
 title = {{OpenAlex}},
 url = {https://openalex.org/},
 key = {OpenAlex},
 note = {Last accessed 2026/04/27}
}

@misc{ClinicalTrialsgov,
 title = {{ClinicalTrials.gov API}},
 url = {https://clinicaltrials.gov/data-api/api},
 key = {ClinicalTrials.gov},
 note = {Last accessed 2026/04/30}
}

@misc{whoictrp,
 title = {{WHO International Clinical Trials Registry Platform (ICTRP) }},
 url = {https://trialsearch.who.int/},
 key = {WHO International Clinical Trials Registry Platform},
 note = {Last accessed 2026/04/30}
}

\end{document}